\documentclass[11pt]{article}

\usepackage[final]{acl}

\usepackage{times}
\usepackage{latexsym}

\usepackage[T1]{fontenc}

\usepackage[utf8]{inputenc}

\usepackage{microtype}

\usepackage{inconsolata}

\usepackage{graphicx}

\usepackage{amsmath}
\usepackage{multirow}
\usepackage{booktabs}
\usepackage{color}
\usepackage{hhline}
\usepackage[capitalize, nameinlink, noabbrev]{cleveref}
\usepackage{tcolorbox}
\usepackage{pifont}
\usepackage{subcaption}

\newcommand{\deepscaler}[0]{DeepScaleR-1.5B-Preview}
\newcommand{\method}[0]{Step-level Advantage Selection}
\newcommand{\shortmethod}[0]{SAS}

\title{Stabilizing Efficient Reasoning with Step-Level Advantage Selection}

\author{\textbf{Han Wang\textsuperscript{1}}\thanks{Work done during internship at AMD.}, \textbf{Xiaodong Yu\textsuperscript{2}}, \textbf{Jialian Wu\textsuperscript{2}}, \textbf{Jiang Liu\textsuperscript{2}}, \\ \textbf{Ximeng Sun\textsuperscript{2}, Mohit Bansal\textsuperscript{1}, Zicheng Liu\textsuperscript{2}} \\
\textsuperscript{1}UNC Chapel Hill, \textsuperscript{2}Advanced Micro Devices, Inc. \\
\texttt{hwang@cs.unc.edu}
}

\begin{document}
\maketitle
\begin{abstract}
Large language models (LLMs) achieve strong reasoning performance by allocating substantial computation at inference time, often generating long and verbose reasoning traces. 
While recent work on efficient reasoning reduces this overhead through length-based rewards or pruning, many approaches are post-trained under a much shorter context window than base-model training, a factor whose effect has not been systematically isolated. 
We first show that short-context post-training alone, using standard GRPO without any length-aware objective, already induces substantial reasoning compression—but at the cost of increasingly unstable training dynamics and accuracy degradation.
To address this, we propose \method{} (\shortmethod{}), which operates at the reasoning-step level and assigns a zero advantage to low-confidence steps in correct rollouts and to high-confidence steps in verifier-failed rollouts, where failures often arise from truncation or verifier issues rather than incorrect reasoning. 
Across diverse mathematical and general reasoning benchmarks, \shortmethod{} improves average Pass@1 accuracy by 0.86 points over the strongest length-aware baseline while reducing average reasoning length by 16.3\%, yielding a better accuracy–efficiency trade-off.\footnote{Our code is publicly available at \url{https://github.com/HanNight/SAS}}
\end{abstract}

\begin{figure}[!ht]
    \centering
    \includegraphics[width=0.95\linewidth]{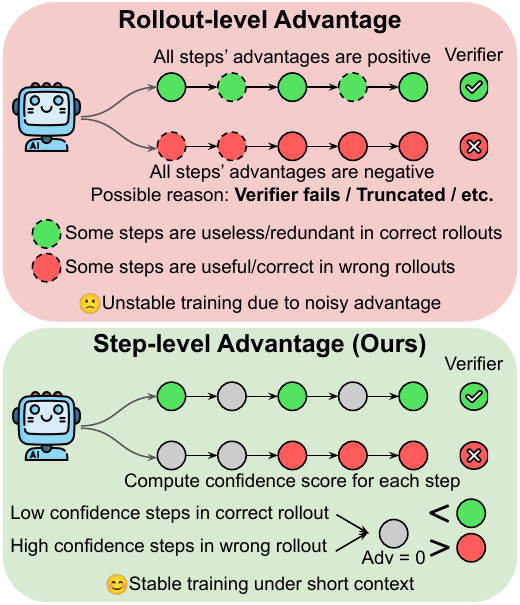}
    \caption{Rollout-level versus step-level advantage selection. Standard GRPO propagates rewards uniformly based on the final verifier outcome: under group-relative normalization, all steps in correct rollouts receive positive advantages and all steps in verifier-failed rollouts receive negative advantages, suppressing useful intermediate reasoning in failed rollouts and over-updating redundant steps in correct rollouts. In contrast, \method{} (\shortmethod{}) assigns a zero advantage to low-confidence steps in correct rollouts (below their positive peers) and to high-confidence steps in verifier-failed rollouts (above their negative peers), resulting in more stable training and efficient reasoning compression under short context.}
    \label{fig:main_fig}
\end{figure}

\section{Introduction}
Large language models (LLMs) have demonstrated strong reasoning capabilities across a wide range of tasks, including mathematical problem solving, logical reasoning, and code generation. Recent progress has shown that allocating more computation at inference time—commonly referred to as test-time scaling—can substantially improve reasoning performance by encouraging models to generate longer chain-of-thought traces or explore multiple reasoning paths \citep{wei2022cot,NEURIPS2022_8bb0d291,wang2023selfconsistency,yao2023tot}. However, these gains often come at a significant computational cost, as models tend to produce excessively long and verbose reasoning even for relatively simple problems, leading to increased inference latency and reduced practical efficiency \citep{chen2025do,gema2025inverse,ghosal2025does}.

To address this issue, a growing body of work has focused on efficient reasoning, which aims to reduce reasoning length while preserving task performance. Many recent approaches adopt reinforcement learning–based post-training strategies that explicitly incorporate length-aware objectives, such as token-budget constraints, length-aware rewards, or pruning mechanisms \citep{aggarwal2025l,wu2025lapo,hou2025thinkprune,sui2025stop}. While effective, these approaches share a critical, yet overlooked, training condition: they typically conduct post-training within a substantially restricted context window (e.g., 4K tokens), a sharp departure from the expansive windows (e.g., 16K–24K tokens) used during base model training. This discrepancy raises a fundamental question: to what extent does the observed reasoning compression stem from explicit length-aware objectives, as opposed to being a natural consequence of short-context post-training itself? 

In this work, we systematically isolate this effect by training a long-context reasoning model \citep{deepscaler2025,guo2025deepseek} using pure GRPO with short context, deliberately excluding any length-aware rewards or pruning techniques. 
Our findings reveal that short-context post-training alone serves as a strong and sufficient compression signal, achieving reductions in output length comparable to state-of-the-art efficient reasoning methods—a critical variable previously conflated with explicit length-control objectives.
However, this compression from short-context post-training comes at a cost: as training progresses, task accuracy fluctuates and degrades, and exploration collapses into brittle policy updates.
We hypothesize that this instability stems from truncation within the restricted context window: truncated rollouts receive zero reward despite often containing correct intermediate reasoning, thereby implicitly penalizing correct steps, leading to unstable behavior and degraded task performance. Short-context post-training alone is therefore effective at compression but insufficient for stable, efficient reasoning.

To resolve this tension between aggressive reasoning compression and stable performance preservation, we introduce \method{} (\shortmethod{}) (\cref{fig:main_fig}), which treats reasoning as a sequence of discrete, evaluable steps. As illustrated in \cref{fig:main_fig}, rather than applying a uniform advantage to an entire trace, our method operates at the reasoning-step level to selectively filter out unreliable steps.
Specifically, we assign a zero advantage to low-confidence reasoning steps in correct rollouts, and to high-confidence intermediate steps within verifier-failed rollouts--which may fail due to truncation under short context or verifier issues rather than incorrect reasoning.
Under GRPO's group-relative advantage normalization, assigning zero advantages has asymmetric effects: it sits below the positive advantages of peers in correct rollouts (suppressing unreliable steps) and above the negative advantages of peers in verifier-failed rollouts (shielding reliable steps from penalization).
By focusing the optimization signal on complete, high-confidence reasoning steps, SAS mitigates noisy updates and preserves performance during aggressive reasoning compression under short context.
We evaluate \shortmethod{} on a diverse set of mathematical and general reasoning benchmarks. 
Experimental results demonstrate that \shortmethod{} consistently achieves a superior accuracy–efficiency trade-off compared to existing baselines. Specifically, relative to the base \deepscaler{} model, our method reduces the average output length by over 30\% while simultaneously improving average Pass@1 accuracy by 1.51 points. Compared to length-aware baselines such as LAPO and ThinkPrune, our approach achieves higher accuracy while generating 15\% fewer tokens on average. These improvements are reflected in the Accuracy–Efficiency Score, where our method exceeds all baselines by a clear margin, with an absolute gain of over 0.13 AES on math reasoning compared to the strongest competing approaches.

\begin{figure*}[t]
    \centering
    \begin{subfigure}[t]{0.48\linewidth}
        \centering
        \includegraphics[width=\linewidth]{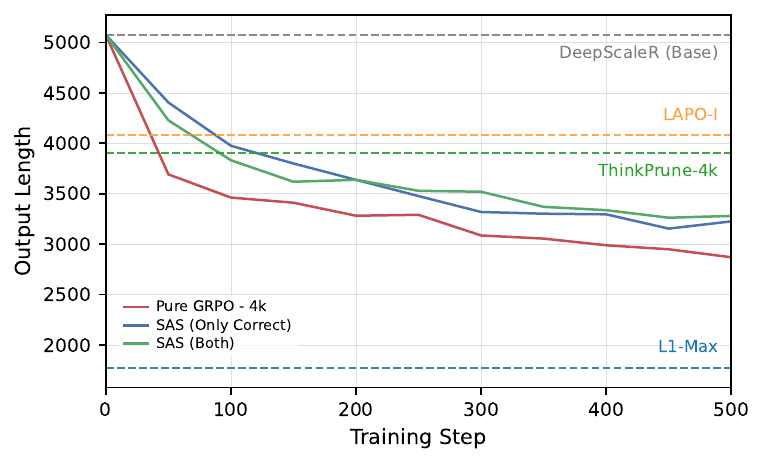}
        \caption{Output length vs.\ training step.}
        \label{fig:len_curve}
    \end{subfigure}
    \hfill
    \begin{subfigure}[t]{0.48\linewidth}
        \centering
        \includegraphics[width=\linewidth]{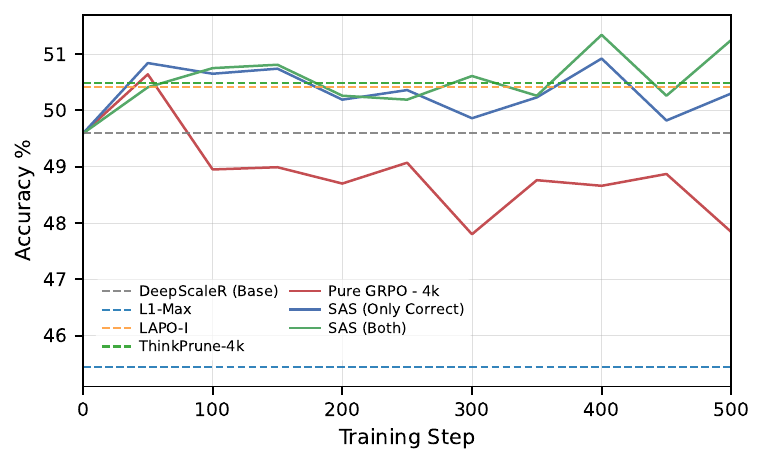}
        \caption{Accuracy vs.\ training step.}
        \label{fig:acc_curve}
    \end{subfigure}

    \caption{Training dynamics of average output length and accuracy across five math reasoning datasets under short-context (4K) post-training. Note that the dashed horizontal lines just indicate the performance of these baselines and do not represent their training trajectories.}
    \label{fig:pure_grpo_4k}
\end{figure*}

\section{Methodology}

\subsection{Preliminary: RL with Length-Aware Reward Design}

Recent work on efficient reasoning commonly adopts reinforcement learning (RL) to explicitly control the length of model-generated reasoning. In this setting, a pretrained language model is further optimized using policy optimization, where the training objective combines task-level rewards with regularization terms that constrain deviations from a reference policy. A widely used framework is Group Relative Policy Optimization (GRPO; \citealp{shao2024deepseekmath}), which optimizes the policy based on relative advantages computed within groups of sampled responses.

Formally, for each prompt $x \sim \mathcal{D}$, GRPO samples a group of $G$ rollouts $\{y_i\}_{i=1}^{G}$ from the current policy $\pi_\theta$ and computes a group-relative advantage for each rollout by normalizing its reward against the group's reward statistics:
\begin{equation}
\hat{A}_{i,t} = \frac{r(x, y_i) - \text{mean}\big(\{r(x, y_j)\}_{j=1}^{G}\big)}{\text{std}\big(\{r(x, y_j)\}_{j=1}^{G}\big)},
\label{eq:advantage}
\end{equation}
which is shared across all tokens $t \in \{1, \ldots, |y_i|\}$ within the same rollout (i.e., $\hat{A}_{i,t} = \hat{A}_i$) under the outcome-based reward $r(x, y_i) \in \{0, 1\}$. The policy is then optimized with a PPO-style clipped surrogate over these advantages, regularized by a KL penalty against a fixed reference policy $\pi_{\text{ref}}$. Crucially, this group-relative normalization ensures that correct rollouts (reward $=1$) receive \textbf{positive} advantages and verifier-failed rollouts (reward $=0$) receive \textbf{negative} advantages.

To encourage concise reasoning, many prior methods incorporate length-aware reward design into the RL framework. A generic form of such rewards can be written as:
\begin{equation}
r(x, y) = r_{\text{task}}(x, y) - \lambda \cdot g(|y|),
\label{eq:length-reward}
\end{equation}
where $r_{\text{task}}(x, y)$ measures task correctness, $|y|$ denotes the output length, and $g(\cdot)$ is a monotonic function that penalizes longer reasoning traces. Different methods instantiate $g(\cdot)$ using target lengths, token budgets, or length-dependent penalties, and integrate this reward into the advantage estimation during policy optimization. Empirically, these approaches show that reasoning length can be directly influenced through RL-based post-training.

However, several empirical observations complicate this formulation. First, prior work reports that output length often exhibits non-monotonic behavior during RL training, where reasoning length initially increases before decreasing later. Second, most length-control methods perform post-training using a substantially shorter context window (e.g., 4K tokens) compared to the long context used during base-model training (e.g., 16--24K). This context mismatch suggests that models trained under shorter contexts may naturally prefer shorter outputs, independent of the specific length-aware reward design. Together, these factors make it unclear to what extent reasoning compression should be attributed to explicit length rewards versus training dynamics and context length.

\subsection{Short-Context Post-Training}
\label{sec:sec3_2}
To isolate the influence of training context length from explicit length-aware reward design, we conduct a controlled study using pure GRPO without any length-dependent reward or constraint. Starting from a reasoning-capable base model pretrained with a long context window, we perform GRPO-based post-training using a fixed 4K context window, which matches the post-training setup commonly adopted in prior efficient reasoning work. Importantly, our reward function is based solely on task correctness, introducing no explicit signal regarding output length.

Despite the absence of length-aware objectives, short-context post-training alone induces a substantial reduction in reasoning length. As shown in~\cref{fig:len_curve}, the average output length decreases sharply during the early stages of training and continues to decline steadily thereafter, eventually reaching a level comparable to or even shorter than existing efficient reasoning baselines such as LAPO and ThinkPrune. At the same time, the trained policy achieves task accuracy that is initially comparable to these methods, as shown in~\cref{fig:acc_curve}. These results indicate that short-context post-training itself provides a strong and previously underexamined compression signal, independent of explicit length-control mechanisms. 

However, a closer inspection of the training dynamics reveals important limitations. While the reasoning length continues to decrease throughout training, task accuracy becomes increasingly volatile, characterized by noticeable fluctuations and a gradual performance decay in later training stages (\cref{fig:acc_curve}). We hypothesize that this instability is tightly coupled to how the short context window interacts with rollout-level credit assignment: when rollouts are truncated before reaching the final answer, the verifier assigns zero reward to traces that may contain largely correct intermediate reasoning, producing noisy and misaligned advantage signals. To quantify how often this occurs, we take 8K-length rollouts from the base model, truncate them to 4K tokens, and re-run the same rule-based verifier. Approximately 29\% of originally correct responses become verifier-failed after truncation, with the large majority losing only the final boxed answer or the closing steps of an otherwise correct derivation. This indicates that a substantial fraction of zero-reward rollouts under short-context training are not logically flawed but merely incomplete, and that pure GRPO systematically penalizes their intermediate reasoning — a plausible driver of the observed accuracy degradation.

Taken together, these observations highlight a fundamental tension between reasoning compression and performance stability: short-context post-training is effective at reducing output length, but the very mechanism that produces this compression (truncation) also injects noisy credit into standard RL updates. This motivates a training strategy that retains the compression benefits of short-context post-training while explicitly correcting for truncation-induced credit misassignment — which we develop next.

\begin{table*}[]
\centering
\resizebox{\textwidth}{!}{
\begin{tabular}{lccccccccccccc}
\toprule
 & \multicolumn{2}{c}{\textbf{AIME24}} & \multicolumn{2}{c}{\textbf{AIME25}} & \multicolumn{2}{c}{\textbf{MATH}} & \multicolumn{2}{c}{\textbf{AMC}} & \multicolumn{2}{c}{\textbf{OlympiadBench}} & \multicolumn{3}{c}{\textbf{Average}} \\ 
 \cmidrule(lr){2-3} \cmidrule(lr){4-5} \cmidrule(lr){6-7} \cmidrule(lr){8-9} \cmidrule(lr){10-11} \cmidrule(lr){12-14}
 & Pass@1 & \#Tok & Pass@1 & \#Tok & Pass@1 & \#Tok & Pass@1 & \#Tok & Pass@1 & \#Tok & Pass@1 & \#Tok & AES \\ \midrule
DeepScaleR & 33.75 & 6755 & 26.88 & 6444 & 86.36 & 2809 & 67.62 & 4761 & 47.23 & 4824 & 52.37 & 5118 & 0.00 \\
GRPO-4K & 38.75 & 5282 & 25.42 & 4812 & 85.09 & 1976 & \textbf{71.69} & 3395 & 47.09 & 3411 & 53.61 & 3775 & 0.33 \\
L1-Max & 25.63 & 2158 & 21.67 & 2032 & 83.60 & 1480 & 64.61 & 1768 & 44.69 & 1703 & 48.04 & 1828 & 0.23 \\ 
LAPO-I & 34.58 & 5627 & \textbf{27.92} & 5290 & 86.04 & 2220 & 69.73 & 3765 & 48.18 & 3735 & 53.29 & 4127 & 0.25 \\
ThinkPrune-4k & 36.04 & 5468 & 26.67 & 5177 & 86.23 & 2090 & 70.18 & 3636 & 47.62 & 3651 & 53.35 & 4004 & 0.27 \\
SAS (ours) & \textbf{39.79} & 4876 & 26.67 & 4295 & \textbf{86.58} & 1768 & 71.46 & 3090 & \textbf{48.19} & 3008 & \textbf{54.54} & 3407 & \textbf{0.46} \\
\bottomrule
\end{tabular}
}
\caption{Comparison of methods across five math reasoning benchmarks. \shortmethod{} achieves a superior accuracy–efficiency trade-off, reducing reasoning length while maintaining or improving accuracy over strong baselines.}
\label{tab:main_res_math}
\end{table*}

\begin{table*}[t]
\small
\centering
\begin{tabular}{lccccccccc}
\toprule
 & \multicolumn{2}{c}{\textbf{GPQA-Diamond}} & \multicolumn{2}{c}{\textbf{LSAT}} & \multicolumn{2}{c}{\textbf{MMLU}} & \multicolumn{3}{c}{\textbf{Average}}  \\
 \cmidrule(lr){2-3} \cmidrule(lr){4-5} \cmidrule(lr){6-7} \cmidrule(lr){8-10}
 & Pass@1 & \#Tok & Pass@1 & \#Tok & Pass@1 & \#Tok & Pass@1 & \#Tok & AES \\ \midrule
 DeepScaleR & 35.70 & 4860 & \textbf{28.64} & 6934 & 47.99 & 1454 & 37.44 & 4416 & 0.00 \\
 GRPO-4K & 32.48 & 1515 & 28.45 & 5025 & 48.73 & 950 & 36.55 & 2496 & 0.32 \\
 L1-Max & 36.05 & 3878 & 26.66 & 1949 & 48.94 & 900 & 37.22 & 2242 & \textbf{0.46} \\
 LAPO-I & 36.17 & 3404 & 28.42 & 5382 & 48.71 & 1207 & 37.77 & 3331 & 0.27 \\
 ThinkPrune-4k & 35.83 & 3278 & 28.13 & 5131 & 48.91 & 973 & 37.62 & 3127 & 0.31 \\
 SAS (ours) & \textbf{37.18} & 2998 & 28.32 & 4312 & \textbf{49.39} & 876 & \textbf{38.30} & 2729 & 0.45 \\
 \bottomrule
\end{tabular}
\caption{Performance comparison of methods on three out-of-domain general reasoning benchmarks. \shortmethod{} consistently improves Pass@1 accuracy while substantially reducing reasoning length, resulting in a strong and competitive accuracy–efficiency trade-off under short-context post-training.}
\label{tab:main_res_general}
\end{table*}

\subsection{\method{} (\shortmethod{})}
\label{sec:sas}

Motivated by the training instability observed during the short-context post-training (Section~\ref{sec:sec3_2}), we propose Step-level Advantage Selection (SAS). The method stabilizes the learning process by selectively modulating the influence of individual reasoning steps on policy updates, without requiring explicit length-aware rewards or architectural changes. SAS operates directly at the advantage level, ensuring that only reliable reasoning steps contribute to the optimization signal.

\paragraph{Mask Advantages in Correct Rollouts.}
Even when a rollout is verified as correct (reward $=1$), the reasoning trace may contain redundant or low-confidence steps (e.g., self-doubt detours, repetitive verification) that are weakly related to the outcome. Reinforcing these steps can introduce noisy updates and exacerbate policy drift during training.

For each correct rollout $y_i$, we partition the generated reasoning trace into a sequence of reasoning steps $\{s_j\}_{j=1}^{N}$, where steps are defined as contiguous text segments separated by double newline delimiters (\texttt{\textbackslash n\textbackslash n})\footnote{See Appendix~\ref{app:step_seg} for a detailed justification and discussion of using \texttt{\textbackslash n\textbackslash n} as the step delimiter.}. For each step $s_j$, we compute a step-level confidence score based on the model's token-level log probabilities. Let $\mathcal{T}_j$ denote the set of token positions belonging to step $s_j$. The confidence score is defined as:
\begin{equation}
c_j = \frac{1}{|\mathcal{T}_j|} \sum_{\tau \in \mathcal{T}_j} \log \pi_\theta\big(y_\tau \mid x, y_{<\tau}\big),
\label{eq:confidence}
\end{equation}
where $\pi_\theta$ denotes the current policy.

Given the confidence scores $\{c_j\}$, we sort the reasoning steps in ascending order of confidence and select a ratio $r \in (0, 1)$ of the lowest-confidence steps. For all token positions belonging to these steps, we set their advantages to zero:
\begin{equation}
\tilde{A}_{i,\tau} = 
\begin{cases}
0, & \text{if } \tau \in \mathcal{T}_j \text{ for } s_j \in \mathcal{S}^+_{\text{mask}}, \\
\hat{A}_{i,\tau}, & \text{otherwise},
\end{cases}
\label{eq:mask-correct}
\end{equation}
where $\hat{A}_{i,\tau}$ is the original GRPO advantage from Eq.~\ref{eq:advantage}, and $\mathcal{S}^+_{\text{mask}}$ denotes the subset of low-confidence steps selected from a verifier-approved rollout. This operation suppresses the contribution of unreliable reasoning steps while preserving learning signals from more reliable ones: the zeroed advantages lie strictly below the positive advantages retained by the remaining high-confidence steps in the same rollout.

\paragraph{Shielding Signals in Incorrect Rollouts.}
We further extend this idea to rollouts that fail the rule-based verification (reward $=0$). A key observation in short-context training is that many ``incorrect'' rollouts are not inherently flawed in logic; rather, they are naturally truncated ``correct'' traces where the model ran out of context before reaching the final answer (Section~\ref{sec:sec3_2}). These traces often contain high-quality intermediate reasoning steps that are identical to those found in successful solutions. Rather than discarding these traces entirely—which would propagate unwarranted negative credit through all of their steps—we shield their reliable segments from penalization.

For each verifier-rejected rollout, we again compute step-level confidence scores $\{c_j\}$ and sort the steps in descending order of confidence. We then select a ratio $r$ of the highest-confidence steps and set their advantages to zero:
\begin{equation}
\tilde{A}_{i,\tau} = 
\begin{cases}
0, & \text{if } \tau \in \mathcal{T}_j \text{ for } s_j \in \mathcal{S}^-_{\text{mask}}, \\
\hat{A}_{i,\tau}, & \text{otherwise},
\end{cases}
\label{eq:mask-failed}
\end{equation}
where $\mathcal{S}^-_{\text{mask}}$ denotes the subset of high-confidence steps selected from a verifier-rejected rollout. The remaining steps retain their original negative advantages, so the zeroed steps sit strictly above them, protecting reliable intermediate reasoning from undue penalization. At the same time, zero lies at or below the positive advantages retained by any step in a correct rollout in the same group, so this shielding never over-rewards steps from failed rollouts relative to legitimately correct reasoning. The result is a denser but well-calibrated learning signal, which is critical for maintaining accuracy during aggressive reasoning compression.

Together, these two components form a reward-conditioned, step-level advantage selection mechanism that reuses a single zero-valued operation. Correct rollouts are prevented from reinforcing unreliable reasoning steps, while verifier-rejected rollouts are decomposed to shield useful intermediate reasoning from undue penalization. As shown in our experiments, this symmetric treatment stabilizes training and improves accuracy–efficiency trade-offs under short-context post-training.

\section{Experiments}

\subsection{Experimental Setup}

\paragraph{Training Details}
We use DeepScaleR-Preview-Dataset \citep{deepscaler2025} for training, which includes approximately 40K mathematics problem–answer pairs collected from AIME (1984–2023), AMC (prior to 2023), Omni-MATH \citep{gao2025omnimath}, and Still \citep{min2024imitate}.
Our base model is \deepscaler{} \citep{deepscaler2025}, a 1.5B-parameter model originally RL fine-tuned from DeepSeek-R1-Distill-Qwen-1.5B \citep{guo2025deepseek} on the same dataset using three training stages with increasing context lengths (8K $\rightarrow$ 16K $\rightarrow$ 24K).
For GRPO training, we adopt the same hyperparameters as in \deepscaler{}. 
In particular, we use a learning rate of 1e-6 and a training batch size of 128.
For our method, we set the hyperparameter ratio $r$ to 0.3.
For a fair comparison, we set the maximum context length as 4K tokens during training.
All experiments are conducted for 500 training steps using the VeRL framework \citep{sheng2025verl}, with 8 rollouts sampled per prompt.
We use AIME24 as validation set to select the checkpoint with the highest Accuracy–Efficiency Score (AES).
All experiments are conducted using 8 AMD MI250 with 64 GB memory.

\paragraph{Evaluation Details}
We evaluate on five different math reasoning datasets: AIME2024, AIME2025, AMC \citep{amc}, MATH \cite{dan2021math}, OlympiadBench \citep{he-etal-2024-olympiadbench}.
In addition, we include GPQA-Diamond \citep{rein2024gpqa}, LSAT \citep{zhong-etal-2024-agieval}, MMLU \citep{hendrycks2021measuring}, three general reasoning benchmarks to test the ability to generalize to out-of-domain data.
For each problem, we sample $k = 16$ responses with a temperature of 0.6, a top-p of 0.95, and a maximum generation length of 8K tokens.
Following standard practices \citep{guo2025deepseek}, we report $\text{Pass@1} = \frac{1}{k}\sum_{i=1}^{k} p_i$, where $p_i \in \{0, 1\}$ denotes the correctness of the i-th response\footnote{This is equivalent to generating $k$ outputs with different random seeds (one output per seed) and averaging their correctness, which reduces evaluation variance compared to relying on a single sampled output.}.
We also report the average number of output tokens to measure reasoning length.
Finally, we compute the Accuracy-Efficiency Score (AES, see details in \cref{app:aes}; \citealp{luo2025opruner}) to evaluate the trade-off between improving accuracy and reducing computational cost.

\paragraph{Baselines}
We evaluate our method against the following baselines, all of which are initialized from \deepscaler{} post-trained using a maximum context length of 4K tokens: 
\begin{itemize}
    \item \textbf{GRPO-4K}: trained with standard GRPO post-training under a 4K training context window, without any additional RL techniques.
    \item \textbf{L1-Max} \citep{aggarwal2025l}: trained with Length Controlled Policy Optimization (LCPO) to generate outputs of varying lengths while respecting a specified maximum length constraint.
    \item \textbf{ThinkPrune-4k} \citep{hou2025thinkprune}: An RL–based pruning method that iteratively enforces progressively stricter token limits to remove redundant reasoning steps.
    \item \textbf{LAPO-I} \citep{wu2025lapo}: A two-stage RL approach that adaptively controls reasoning length by modeling and leveraging successful solution length distributions.
\end{itemize}

\subsection{Experimental Results}
\cref{tab:main_res_math} presents the performance of SAS and baselines across five mathematical reasoning datasets. Our method achieves the best accuracy–efficiency trade-off, outperforming all baselines.
Notably, short-context post-training alone already induces substantial reasoning compression: compared to the base \deepscaler{}, GRPO-4K substantially reduces the average output length while slightly improving average Pass@1 accuracy, confirming that context length itself provides a strong compression signal. However, this compression comes with limited stability (as shown in \cref{fig:pure_grpo_4k}).
Among length-aware baselines, L1-Max achieves the most aggressive compression but at a significant cost to task accuracy. Conversely, ThinkPrune and LAPO prioritize accuracy preservation but achieve only moderate length reduction, resulting in lower overall efficiency gains. In contrast, SAS consistently outperforms all the baselines, improving average Pass@1 accuracy by more than 2 points over the base model while reducing output length by approximately 1,700 tokens. This dual improvement results in the highest AES of 0.46. Notably, our method maintains or exceeds the accuracy of GRPO-4K across all math benchmarks while generating shorter traces, demonstrating a better accuracy–efficiency trade-off.

We extend our evaluation to three general reasoning benchmarks in \cref{tab:main_res_general}. Although these tasks typically require more concise reasoning traces, we observe similar trends. Pure short-context post-training (GRPO-4K) again leads to reduced output length but degrades accuracy relative to the base model, indicating over-compression. In contrast, our method maintains or improves accuracy across all general reasoning datasets with substantially shorter outputs, achieving the best overall efficiency score among stable methods. These results demonstrate that the benefits of our approach are not limited to mathematical reasoning, but extend to broader reasoning settings where controlling training stability is critical.

\section{Analysis}

\begin{figure}
    \centering
    \includegraphics[width=1\linewidth]{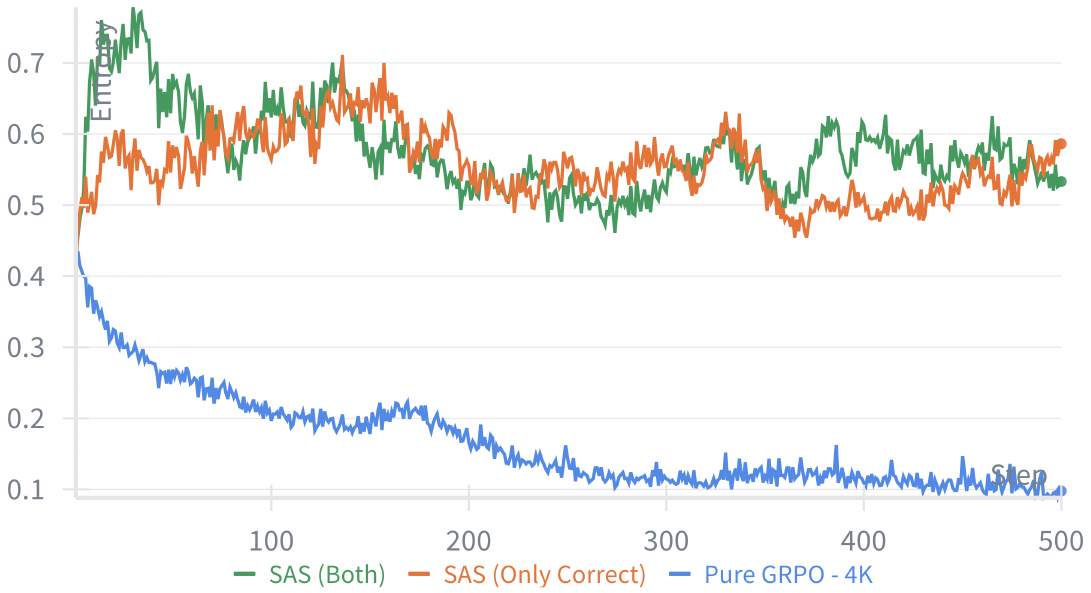}
    \caption{Policy entropy throughout training. SAS maintains higher entropy compared to the rapid entropy collapse observed in pure GRPO-4K, indicating more robust exploration.}
    \label{fig:entropy}
\end{figure}

\begin{table*}[t]
\centering
\resizebox{\textwidth}{!}{
\begin{tabular}{lccccccccccccc}
\toprule
 & \multicolumn{2}{c}{\textbf{AIME24}} & \multicolumn{2}{c}{\textbf{AIME25}} & \multicolumn{2}{c}{\textbf{MATH}} & \multicolumn{2}{c}{\textbf{AMC}} & \multicolumn{2}{c}{\textbf{OlympiadBench}} & \multicolumn{3}{c}{\textbf{Average}} \\ 
 \cmidrule(lr){2-3} \cmidrule(lr){4-5} \cmidrule(lr){6-7} \cmidrule(lr){8-9} \cmidrule(lr){10-11} \cmidrule(lr){12-14}
  & Pass@1 & \#Tok & Pass@1 & \#Tok & Pass@1 & \#Tok & Pass@1 & \#Tok & Pass@1 & \#Tok & Pass@1 & \#Tok & AES \\ \midrule
DeepScaleR & 33.75 & 6755 & \textbf{26.88} & 6444 & 86.36 & 2809 & 67.62 & 4761 & 47.23 & 4824 & 52.37 & 5118 & 0.00 \\
GRPO-4K & 38.75 & 5282 & 25.42 & 4812 & 85.09 & 1976 & 71.69 & 3395 & 47.09 & 3411 & 53.61 & 3775 & 0.33 \\
SAS (ours) & \textbf{39.79} & 4876 & 26.67 & 4295 & \textbf{86.58} & 1768 & 71.46 & 3090 & \textbf{48.19} & 3008 & \textbf{54.54} & 3407 & \textbf{0.46} \\
~~- Only Correct & 36.46 & 4686 & \textbf{26.88} & 4294 & 86.18 & 1801 & 71.99 & 3082 & 47.99 & 2965 & 53.90 & 3366 & 0.43 \\
~~- Random Steps & 37.29 & 4892 & 24.79 & 4461 & 86.10 & 1780 & 71.16 & 3086 & 47.51 & 3055 & 53.37 & 3455 & 0.38 \\
~~- Token Level & 36.25 & 4771 & 25.00 & 4295 & 86.11 & 1912 & \textbf{72.36} & 3148 & 47.56 & 3071 & 53.46 & 3439 & 0.39 \\
\bottomrule
\end{tabular}
}
\caption{Experimental results for the ablation study.}
\label{tab:ablation}
\end{table*}

\subsection{Ablation Study}
We evaluate the impact of our design choices on five math reasoning datasets, as shown in \cref{tab:ablation}. 

\paragraph{Value of Verifier-Failed Rollouts.}
We first examine the necessity of leveraging verifier-failed rollouts. Restricting SAS to only correct rollouts (``Only Correct'') reduces the average AES from 0.46 to 0.43, with accuracy dropping from 54.54 to 53.90 despite comparable output length.
This suggests that shielding high-confidence intermediate steps in verifier-failed rollouts from undue penalization is critical for stabilizing the model during aggressive compression: without this mechanism, truncation-induced verifier failures propagate negative credit to reasoning that is actually correct, destabilizing training.
We further analyze the training stability by monitoring the policy entropy, as shown in \cref{fig:entropy}. While the entropy of pure GRPO-4K (blue line) drops rapidly, signaling a collapse in exploration and the adoption of brittle, repetitive reasoning patterns, both SAS (Both) and SAS (Only Correct) maintain a significantly higher and more stable entropy level throughout training.
This suggests that by neutralizing noisy advantage signals---rather than propagating them through all steps of a rollout---SAS prevents the model from converging prematurely to suboptimal reasoning patterns, thereby preserving the diversity and robustness of the reasoning process.

\paragraph{Confidence-Aware Step Selection.}
Replacing our confidence-based selection with random step selection (``Random Steps'') results in a clear degradation in efficiency (AES 0.38) and longer reasoning traces. This indicates that the gains of our method do not arise merely from sparsifying the advantage signal, but critically depend on selectively filtering low-confidence reasoning steps.

\paragraph{Step-Level vs. Token-Level Granularity.}
The token-level variant (``Token Level'') underperforms our step-level design, achieving lower average accuracy and AES (0.39 vs. 0.46), while producing longer outputs. This confirms that aggregating tokens into semantically meaningful reasoning steps yields more stable and effective advantage selection than fine-grained token-level selection.

Overall, these ablations demonstrate that all components of our method—leveraging both correct and incorrect rollouts, step-level granularity, and confidence-based selection—jointly contribute to the strongest accuracy–efficiency trade-off under short-context post-training.

\subsection{Effect of the Selection Ratio}

We study the effect of the selection ratio \(r\) in Step-level Advantage Selection (SAS), which controls the fraction of reasoning steps selected for advantage assignment during training. We vary \(r\) from 0.1 to 0.9 while keeping all other training settings fixed, and compare against both the base \deepscaler{} model and pure GRPO-4K.
As shown in \cref{tab:selection_ratio}, SAS consistently outperforms the base \deepscaler{} model across all tested ratios, yielding higher accuracy and substantially shorter reasoning traces. The best overall performance is achieved at \(r = 0.3\), which attains the highest Pass@1 accuracy (54.54), the strongest accuracy–efficiency trade-off (AES = 0.46), and a large reduction in output length compared to the base model. However, performance differences across ratios are relatively small: even extreme values $r = 0.9$ remain competitive, with AES scores above 0.36 and stable accuracy.
One possible explanation is that, even when most steps are retained, \shortmethod{} still induces a highly non-uniform learning signal in which only a small subset of informative reasoning steps meaningfully influences policy updates. This aligns with recent findings that reinforcement learning for LLM reasoning is primarily driven by a minority of high-entropy or decision-critical steps, while most tokens in long reasoning traces are effectively redundant from an optimization perspective \citep{wang2025beyond}.
Overall, these results show that \shortmethod{} is practically robust and easy to deploy. While moderate ratios offer the best accuracy–efficiency trade-off, the method performs reliably across a wide range of settings. This reinforces the central conclusion that \emph{which} reasoning steps are selected is more important than \emph{how many}, highlighting step-level advantage selection as the key driver of stable and effective reasoning compression.

\begin{table}[]
\small
\centering
\begin{tabular}{cccc}
\toprule
Selection Ratio ($r$) & Pass@1 & \#Tok & AES \\ \midrule
DeepScaleR & 52.37 & 5118 & 0.00 \\
GRPO-4K & 53.61 & 3775 & 0.33 \\ \midrule \midrule
0.1 & 53.52 & 3259 & 0.43 \\
0.3 & \textbf{54.54} & 3407 & \textbf{0.46} \\
0.5 & 53.39 & 3407 & 0.39 \\
0.7 & 53.22 & 3412 & 0.38 \\
0.9 & 53.06 & 3482 & 0.36 \\
\bottomrule
\end{tabular}
\caption{Average performance of \shortmethod{} under different selection ratios $r$ on five math reasoning datasets.}
\label{tab:selection_ratio}
\end{table}

\subsection{Validating Step Confidence}
A natural question is whether the policy's own token log-probabilities faithfully reflect step quality. 
To validate this, we sample 16 responses per question on MATH500 from \deepscaler{} (8,000 responses total), segment each response into reasoning steps using double newlines (\texttt{\textbackslash n\textbackslash n}), and score every step independently with (i) the mean of token-level log probabilities within that step, and (ii) an external Process Reward Model, Qwen2.5-Math-PRM-7B \citep{zhang-etal-2025-lessons}.
We then measure the ranking correlation between these two scores using nDCG@k.
The resulting correlation is 0.9022, indicating strong agreement between our confidence-based ranking and the PRM-based ranking.
This suggests that token-level log probabilities provide a reliable signal for identifying high-quality reasoning steps.
Moreover, introducing a PRM can lead to reward hacking \citep{pmlr-v202-gao23h} and significantly increase computational overhead during large-scale RL training \citep{guo2025deepseek}. In contrast, SAS does not rely on an auxiliary reward model and leverages intrinsic confidence signals already available from the policy, avoiding additional training complexity and cost. Taken together, our empirical correlation analysis and the practical considerations support the use of token-level log probabilities as an effective and scalable proxy for step-level confidence in SAS.

\subsection{Computational Overhead of SAS}
We measured the computational overhead of SAS under the same training configuration (training batch size = 128) as standard GRPO. In this setting, the average training time per step is 279.08 seconds for standard GRPO and 327.15 seconds for SAS. This corresponds to an approximately 17\% increase in per-step wall-clock time. The additional cost mainly comes from step segmentation and step-level advantage computation, which introduce lightweight post-processing over generated tokens but do not require additional forward passes, auxiliary models, or extra rollouts. Importantly, SAS does not modify the model architecture or the sampling procedure, and the memory footprint remains unchanged. Given that SAS consistently improves the accuracy–efficiency trade-off while introducing only moderate computational overhead, we believe this cost is practical and acceptable.

\section{Related Work}
\paragraph{Test-time Scaling in LLMs}
Test-time scaling has been shown to improve the performance of large language models on various complex reasoning tasks, such as mathematical problem-solving and code generation \citep{wei2022cot,NEURIPS2022_8bb0d291,wu2025inference}.
Such performance gains can often be achieved by allocating more inference-time computation, either through generating longer chain-of-thought reasoning traces \citep{aman2023selfrefine} or by exploring a larger number of reasoning paths \citep{wang2023selfconsistency,yao2023tot,wang-etal-2024-soft}.
More recently, reinforcement learning has been used to directly induce extended reasoning behaviors, producing models that generate substantially longer and more elaborate reasoning traces at inference time, such as OpenAI-o1 \citep{jaech2024openai} and DeepSeek-R1 \citep{guo2025deepseek}. While these approaches have demonstrated strong gains on challenging benchmarks, they often rely on generating significantly longer reasoning traces, which can be several times longer than those produced by short chain-of-thought models, leading to increased inference cost and latency. Subsequent analyses have found that such extended reasoning frequently contains redundant or unnecessary steps, including excessive verification or repetition, even on relatively simple problems, resulting in an ``overthinking'' phenomenon that degrades efficiency without proportional accuracy gains \citep{chen2025do,ghosal2025does,gema2025inverse}. As a result, despite their effectiveness, existing test-time scaling methods generally lack precise and dynamic control over the length of generated reasoning, motivating growing interest in approaches that retain the benefits of increased inference-time computation while enabling more efficient and controllable reasoning.

\paragraph{Efficient Reasoning for LLMs}
Motivated by the high computational cost and inefficiency introduced by test-time scaling, a growing body of work has focused on \emph{efficient reasoning}, which aims to reduce reasoning overhead while preserving task performance. Rather than allocating additional computation at inference time, these approaches seek to shorten reasoning traces through training-time optimization, reasoning structure compression, or inference control \cite{sui2025stop}.
A prominent class of methods focuses on model-based efficient reasoning, particularly through leveraging traditional RL optimization techniques combined with explicit length-aware reward to control the length of CoT reasoning \citep{team2025kimi,arora2025training}. Representative approaches include L1 \citep{aggarwal2025l}, which introduces explicit length constraints during policy optimization, length-adaptive policy optimization methods such as LAPO \citep{wu2025lapo}, and pruning-based strategies such as ThinkPrune \citep{hou2025thinkprune}, which iteratively remove redundant reasoning steps while maintaining correctness.
Despite their effectiveness, most existing methods entangle explicit length-aware objectives with short-context post-training, making it difficult to isolate the role of training context and learning signal design. In contrast, our work revisits efficient reasoning from the perspective of credit assignment, showing that stable reasoning compression can be achieved without explicit length rewards by selectively assigning advantages at the level of reasoning steps.

\paragraph{Confidence-based and Entropy-based RL.}
A parallel line of work leverages confidence or entropy signals during RL: \citet{prabhudesai2025maximizing} replaces the verifier-based reward with a negative-entropy reward to directly optimize for high-confidence rollouts, while \citet{wang2025beyond} updates only the highest-entropy tokens and reports \emph{longer} resulting responses. SAS neither modifies the reward nor introduces entropy regularization; confidence serves only as a criterion for selecting which steps receive a nonzero advantage. Furthermore, unlike \citet{wang2025beyond}, SAS \emph{filters} low-confidence steps to mitigate overthinking under short-context post-training, rather than amplifying high-entropy ones.

\section{Discussion and Conclusion}
We revisit efficient reasoning through the lens of post-training context length and advantage selection. We show that short-context post-training alone is sufficient to induce substantial reasoning compression in long-context reasoning models, but our analysis also reveals a previously overlooked limitation of rollout-level reinforcement learning: truncated or verifier-failed rollouts introduce noisy and misaligned learning signals that undermine training stability. These findings highlight that reasoning efficiency is not only shaped by reward design, but also by the granularity at which credit is assigned within a rollout.
To address this, we propose \method{} (\shortmethod{}), which refines advantage assignment through a single zero-valued operation with asymmetric effects: suppressing low-confidence steps in correct rollouts and shielding high-confidence intermediate reasoning in verifier-failed rollouts from undue penalization. \shortmethod{} stabilizes training and achieves a stronger accuracy–efficiency trade-off than existing length-aware methods, demonstrating that stable and effective reasoning compression can be achieved with minimal modifications to standard RL pipelines.

\section*{Limitations}
Our proposed method \method{} (\shortmethod{}) demonstrates strong empirical results on both in-domain and out-of-domain tasks. However, our experiments focus on a single base model, and it remains to be seen how well \shortmethod{} generalizes to models with different sizes, pretraining or post-training paradigms. In addition, all experiments are conducted under a fixed short-context post-training setting, and the behavior of \shortmethod{} across varying training context length has not been systematically studied. Further, while our ablation results suggest that the selective advantage strategy is broadly effective, we leave a deeper theoretical understanding of advantage selection in reasoning language models to future work.
We do not foresee any particular risks associated with the application of our method.

\section*{Acknowledgement}
We thank the anonymous reviewers for their valuable feedback.

\bibliography{custom}

@inproceedings{aggarwal2025l,
title={L1: Controlling How Long A Reasoning Model Thinks With Reinforcement Learning},
author={Pranjal Aggarwal and Sean Welleck},
booktitle={Second Conference on Language Modeling},
year={2025},
}

@article{hou2025thinkprune,
  title={ThinkPrune: Pruning Long Chain-of-Thought of LLMs via Reinforcement Learning}, 
  author={Bairu Hou and Yang Zhang and Jiabao Ji and Yujian Liu and Kaizhi Qian and Jacob Andreas and Shiyu Chang},
  journal={arXiv preprint arXiv:2504.01296},
  year={2025}
}

@inproceedings{he-etal-2024-olympiadbench,
    title = "{O}lympiad{B}ench: A Challenging Benchmark for Promoting {AGI} with Olympiad-Level Bilingual Multimodal Scientific Problems",
    author = "He, Chaoqun  and
      Luo, Renjie  and
      Bai, Yuzhuo  and
      Hu, Shengding  and
      Thai, Zhen  and
      Shen, Junhao  and
      Hu, Jinyi  and
      Han, Xu  and
      Huang, Yujie  and
      Zhang, Yuxiang  and
      Liu, Jie  and
      Qi, Lei  and
      Liu, Zhiyuan  and
      Sun, Maosong",
    editor = "Ku, Lun-Wei  and
      Martins, Andre  and
      Srikumar, Vivek",
    booktitle = "Proceedings of the 62nd Annual Meeting of the Association for Computational Linguistics (Volume 1: Long Papers)",
    month = aug,
    year = "2024",
    address = "Bangkok, Thailand",
    publisher = "Association for Computational Linguistics",
    url = "https://aclanthology.org/2024.acl-long.211/",
    doi = "10.18653/v1/2024.acl-long.211",
    pages = "3828--3850"
}

@inproceedings{rein2024gpqa,
title={{GPQA}: A Graduate-Level Google-Proof Q\&A Benchmark},
author={David Rein and Betty Li Hou and Asa Cooper Stickland and Jackson Petty and Richard Yuanzhe Pang and Julien Dirani and Julian Michael and Samuel R. Bowman},
booktitle={First Conference on Language Modeling},
year={2024},
url={https://openreview.net/forum?id=Ti67584b98}
}

@inproceedings{dan2021math,
 author = {Hendrycks, Dan and Burns, Collin and Kadavath, Saurav and Arora, Akul and Basart, Steven and Tang, Eric and Song, Dawn and Steinhardt, Jacob},
 booktitle = {Proceedings of the Neural Information Processing Systems Track on Datasets and Benchmarks},
 title = {Measuring Mathematical Problem Solving With the MATH Dataset},
 url = {https://datasets-benchmarks-proceedings.neurips.cc/paper_files/paper/2021/file/be83ab3ecd0db773eb2dc1b0a17836a1-Paper-round2.pdf},
 volume = {1},
 year = {2021}
}

@inproceedings{deepscaler2025,
title={DeepScale{R}: Effective {RL} Scaling of Reasoning Models via Iterative Context Lengthening},
author={Michael Luo and Sijun Tan and Justin Wong and Xiaoxiang Shi and William Y. Tang and Manan Roongta and Colin Cai and Jeffrey Luo and Li Erran Li and Raluca Ada Popa and Ion Stoica},
year={2025},
url={https://openreview.net/forum?id=I6GzDCne7U},
note={Notion Blog}
}

@article{shao2024deepseekmath,
  title={Deepseekmath: Pushing the limits of mathematical reasoning in open language models},
  author={Shao, Zhihong and Wang, Peiyi and Zhu, Qihao and Xu, Runxin and Song, Junxiao and Bi, Xiao and Zhang, Haowei and Zhang, Mingchuan and Li, YK and others},
  journal={arXiv preprint arXiv:2402.03300},
  year={2024}
}

@article{guo2025deepseek,
  title={Deepseek-r1: Incentivizing reasoning capability in llms via reinforcement learning},
  author={Guo, Daya and Yang, Dejian and Zhang, Haowei and Song, Junxiao and Zhang, Ruoyu and Xu, Runxin and Zhu, Qihao and Ma, Shirong and Wang, Peiyi and Bi, Xiao and others},
  journal={arXiv preprint arXiv:2501.12948},
  year={2025}
}

@article{wu2025lapo,
  title={Lapo: Internalizing reasoning efficiency via length-adaptive policy optimization},
  author={Wu, Xingyu and Yan, Yuchen and Lyu, Shangke and Wu, Linjuan and Qiu, Yiwen and Shen, Yongliang and Lu, Weiming and Shao, Jian and Xiao, Jun and Zhuang, Yueting},
  journal={arXiv preprint arXiv:2507.15758},
  year={2025}
}

@inproceedings{gao2025omnimath,
title={Omni-{MATH}: A Universal Olympiad Level Mathematic Benchmark for Large Language Models},
author={Bofei Gao and Feifan Song and Zhe Yang and Zefan Cai and Yibo Miao and Qingxiu Dong and Lei Li and Chenghao Ma and Liang Chen and Runxin Xu and Zhengyang Tang and Benyou Wang and Daoguang Zan and Shanghaoran Quan and Ge Zhang and Lei Sha and Yichang Zhang and Xuancheng Ren and Tianyu Liu and Baobao Chang},
booktitle={The Thirteenth International Conference on Learning Representations},
year={2025},
url={https://openreview.net/forum?id=yaqPf0KAlN}
}

@article{min2024imitate,
  title={Imitate, explore, and self-improve: A reproduction report on slow-thinking reasoning systems},
  author={Min, Yingqian and Chen, Zhipeng and Jiang, Jinhao and Chen, Jie and Deng, Jia and Hu, Yiwen and Tang, Yiru and Wang, Jiapeng and Cheng, Xiaoxue and Song, Huatong and others},
  journal={arXiv preprint arXiv:2412.09413},
  year={2024}
}

@misc{amc,
  title        = {American Mathematics Competitions (AMC)},
  author       = {MAA, Mathematical Association of America},
  year         = {2024},
  howpublished = {\url{https://maa.org/student-programs/amc/}},
}

@inproceedings{wei2022cot,
 author = {Wei, Jason and Wang, Xuezhi and Schuurmans, Dale and Bosma, Maarten and ichter, brian and Xia, Fei and Chi, Ed and Le, Quoc V and Zhou, Denny},
 booktitle = {Advances in Neural Information Processing Systems},
 editor = {S. Koyejo and S. Mohamed and A. Agarwal and D. Belgrave and K. Cho and A. Oh},
 pages = {24824--24837},
 title = {Chain-of-Thought Prompting Elicits Reasoning in Large Language Models},
 url = {https://proceedings.neurips.cc/paper_files/paper/2022/file/9d5609613524ecf4f15af0f7b31abca4-Paper-Conference.pdf},
 volume = {35},
 year = {2022}
}

@inproceedings{wang2023selfconsistency,
title={Self-Consistency Improves Chain of Thought Reasoning in Language Models},
author={Xuezhi Wang and Jason Wei and Dale Schuurmans and Quoc V Le and Ed H. Chi and Sharan Narang and Aakanksha Chowdhery and Denny Zhou},
booktitle={The Eleventh International Conference on Learning Representations },
year={2023},
url={https://openreview.net/forum?id=1PL1NIMMrw}
}

@inproceedings{yao2023tot,
 author = {Yao, Shunyu and Yu, Dian and Zhao, Jeffrey and Shafran, Izhak and Griffiths, Tom and Cao, Yuan and Narasimhan, Karthik},
 booktitle = {Advances in Neural Information Processing Systems},
 editor = {A. Oh and T. Naumann and A. Globerson and K. Saenko and M. Hardt and S. Levine},
 pages = {11809--11822},
 publisher = {Curran Associates, Inc.},
 title = {Tree of Thoughts: Deliberate Problem Solving with Large Language Models},
 url = {https://proceedings.neurips.cc/paper_files/paper/2023/file/271db9922b8d1f4dd7aaef84ed5ac703-Paper-Conference.pdf},
 volume = {36},
 year = {2023}
}

@inproceedings{wang-etal-2024-soft,
    title = "Soft Self-Consistency Improves Language Models Agents",
    author = "Wang, Han  and
      Prasad, Archiki  and
      Stengel-Eskin, Elias  and
      Bansal, Mohit",
    editor = "Ku, Lun-Wei  and
      Martins, Andre  and
      Srikumar, Vivek",
    booktitle = "Proceedings of the 62nd Annual Meeting of the Association for Computational Linguistics (Volume 2: Short Papers)",
    month = aug,
    year = "2024",
    address = "Bangkok, Thailand",
    publisher = "Association for Computational Linguistics",
    url = "https://aclanthology.org/2024.acl-short.28/",
    doi = "10.18653/v1/2024.acl-short.28",
    pages = "287--301",
}

@inproceedings{sheng2025verl,
author = {Sheng, Guangming and Zhang, Chi and Ye, Zilingfeng and Wu, Xibin and Zhang, Wang and Zhang, Ru and Peng, Yanghua and Lin, Haibin and Wu, Chuan},
title = {HybridFlow: A Flexible and Efficient RLHF Framework},
year = {2025},
isbn = {9798400711961},
publisher = {Association for Computing Machinery},
address = {New York, NY, USA},
url = {https://doi.org/10.1145/3689031.3696075},
doi = {10.1145/3689031.3696075},
booktitle = {Proceedings of the Twentieth European Conference on Computer Systems},
pages = {1279–1297},
numpages = {19},
series = {EuroSys '25}
}

@inproceedings{luo2025opruner,
title={O1-Pruner: Length-Harmonizing Fine-Tuning for O1-Like Reasoning Pruning},
author={Haotian Luo and Li Shen and Haiying He and Yibo Wang and Shiwei Liu and Wei Li and Naiqiang Tan and Xiaochun Cao and Dacheng Tao},
booktitle={2nd AI for Math Workshop @ ICML 2025},
year={2025},
url={https://openreview.net/forum?id=ioYybCRcyW}
}

@inproceedings{NEURIPS2022_8bb0d291,
 author = {Kojima, Takeshi and Gu, Shixiang (Shane) and Reid, Machel and Matsuo, Yutaka and Iwasawa, Yusuke},
 booktitle = {Advances in Neural Information Processing Systems},
 editor = {S. Koyejo and S. Mohamed and A. Agarwal and D. Belgrave and K. Cho and A. Oh},
 pages = {22199--22213},
 publisher = {Curran Associates, Inc.},
 title = {Large Language Models are Zero-Shot Reasoners},
 url = {https://proceedings.neurips.cc/paper_files/paper/2022/file/8bb0d291acd4acf06ef112099c16f326-Paper-Conference.pdf},
 volume = {35},
 year = {2022}
}

@inproceedings{wu2025inference,
title={Inference Scaling Laws: An Empirical Analysis of Compute-Optimal Inference for {LLM} Problem-Solving},
author={Yangzhen Wu and Zhiqing Sun and Shanda Li and Sean Welleck and Yiming Yang},
booktitle={The Thirteenth International Conference on Learning Representations},
year={2025},
url={https://openreview.net/forum?id=VNckp7JEHn}
}

@inproceedings{aman2023selfrefine,
 author = {Madaan, Aman and Tandon, Niket and Gupta, Prakhar and Hallinan, Skyler and Gao, Luyu and Wiegreffe, Sarah and Alon, Uri and Dziri, Nouha and Prabhumoye, Shrimai and Yang, Yiming and Gupta, Shashank and Majumder, Bodhisattwa Prasad and Hermann, Katherine and Welleck, Sean and Yazdanbakhsh, Amir and Clark, Peter},
 booktitle = {Advances in Neural Information Processing Systems},
 editor = {A. Oh and T. Naumann and A. Globerson and K. Saenko and M. Hardt and S. Levine},
 pages = {46534--46594},
 publisher = {Curran Associates, Inc.},
 title = {Self-Refine: Iterative Refinement with Self-Feedback},
 url = {https://proceedings.neurips.cc/paper_files/paper/2023/file/91edff07232fb1b55a505a9e9f6c0ff3-Paper-Conference.pdf},
 volume = {36},
 year = {2023}
}

@article{jaech2024openai,
  title={Openai o1 system card},
  author={Jaech, Aaron and Kalai, Adam and Lerer, Adam and Richardson, Adam and El-Kishky, Ahmed and Low, Aiden and Helyar, Alec and Madry, Aleksander and Beutel, Alex and Carney, Alex and others},
  journal={arXiv preprint arXiv:2412.16720},
  year={2024}
}

@inproceedings{chen2025do,
title={Do {NOT} Think That Much for 2+3=? On the Overthinking of Long Reasoning Models},
author={Xingyu Chen and Jiahao Xu and Tian Liang and Zhiwei He and Jianhui Pang and Dian Yu and Linfeng Song and Qiuzhi Liu and Mengfei Zhou and Zhuosheng Zhang and Rui Wang and Zhaopeng Tu and Haitao Mi and Dong Yu},
booktitle={Forty-second International Conference on Machine Learning},
year={2025},
url={https://openreview.net/forum?id=MSbU3L7V00}
}

@inproceedings{ghosal2025does,
title={Does Thinking More Always Help? Mirage of Test-Time Scaling in Reasoning Models},
author={Soumya Suvra Ghosal and Souradip Chakraborty and Avinash Reddy and Yifu Lu and Mengdi Wang and Dinesh Manocha and Furong Huang and Mohammad Ghavamzadeh and Amrit Singh Bedi},
booktitle={The Thirty-ninth Annual Conference on Neural Information Processing Systems},
year={2025},
url={https://openreview.net/forum?id=tKPqbamNb9}
}

@article{gema2025inverse,
title={Inverse Scaling in Test-Time Compute},
author={Aryo Pradipta Gema and Alexander H{\"a}gele and Runjin Chen and Andy Arditi and Jacob Goldman-Wetzler and Kit Fraser-Taliente and Henry Sleight and Linda Petrini and Julian Michael and Beatrice Alex and Pasquale Minervini and Yanda Chen and Joe Benton and Ethan Perez},
journal={Transactions on Machine Learning Research},
issn={2835-8856},
year={2025},
url={https://openreview.net/forum?id=NXgyHW1c7M},
note={Featured Certification, J2C Certification}
}

@article{sui2025stop,
title={Stop Overthinking: A Survey on Efficient Reasoning for Large Language Models},
author={Yang Sui and Yu-Neng Chuang and Guanchu Wang and Jiamu Zhang and Tianyi Zhang and Jiayi Yuan and Hongyi Liu and Andrew Wen and Shaochen Zhong and Na Zou and Hanjie Chen and Xia Hu},
journal={Transactions on Machine Learning Research},
issn={2835-8856},
year={2025},
url={https://openreview.net/forum?id=HvoG8SxggZ},
note={}
}

@inproceedings{zhong-etal-2024-agieval,
    title = "{AGIE}val: A Human-Centric Benchmark for Evaluating Foundation Models",
    author = "Zhong, Wanjun  and
      Cui, Ruixiang  and
      Guo, Yiduo  and
      Liang, Yaobo  and
      Lu, Shuai  and
      Wang, Yanlin  and
      Saied, Amin  and
      Chen, Weizhu  and
      Duan, Nan",
    editor = "Duh, Kevin  and
      Gomez, Helena  and
      Bethard, Steven",
    booktitle = "Findings of the Association for Computational Linguistics: NAACL 2024",
    month = jun,
    year = "2024",
    address = "Mexico City, Mexico",
    publisher = "Association for Computational Linguistics",
    url = "https://aclanthology.org/2024.findings-naacl.149/",
    doi = "10.18653/v1/2024.findings-naacl.149",
    pages = "2299--2314"
}

@inproceedings{hendrycks2021measuring,
title={Measuring Massive Multitask Language Understanding},
author={Dan Hendrycks and Collin Burns and Steven Basart and Andy Zou and Mantas Mazeika and Dawn Song and Jacob Steinhardt},
booktitle={International Conference on Learning Representations},
year={2021},
url={https://openreview.net/forum?id=d7KBjmI3GmQ}
}

@inproceedings{arora2025training,
title={Training Language Models to Reason Efficiently},
author={Daman Arora and Andrea Zanette},
booktitle={The Thirty-ninth Annual Conference on Neural Information Processing Systems},
year={2025},
url={https://openreview.net/forum?id=AiZxn84Wdo}
}

@article{team2025kimi,
  title={Kimi k1.5: Scaling reinforcement learning with llms},
  author={Team, Kimi and Du, Angang and Gao, Bofei and Xing, Bowei and Jiang, Changjiu and Chen, Cheng and Li, Cheng and Xiao, Chenjun and Du, Chenzhuang and Liao, Chonghua and others},
  journal={arXiv preprint arXiv:2501.12599},
  year={2025}
}

@inproceedings{wang2025beyond,
title={Beyond the 80/20 Rule: High-Entropy Minority Tokens Drive Effective Reinforcement Learning for {LLM} Reasoning},
author={Shenzhi Wang and Le Yu and Chang Gao and Chujie Zheng and Shixuan Liu and Rui Lu and Kai Dang and Xiong-Hui Chen and Jianxin Yang and Zhenru Zhang and Yuqiong Liu and An Yang and Andrew Zhao and Yang Yue and Shiji Song and Bowen Yu and Gao Huang and Junyang Lin},
booktitle={The Thirty-ninth Annual Conference on Neural Information Processing Systems},
year={2025},
url={https://openreview.net/forum?id=yfcpdY4gMP}
}

@inproceedings{zhang-etal-2025-lessons,
    title = "The Lessons of Developing Process Reward Models in Mathematical Reasoning",
    author = "Zhang, Zhenru  and
      Zheng, Chujie  and
      Wu, Yangzhen  and
      Zhang, Beichen  and
      Lin, Runji  and
      Yu, Bowen  and
      Liu, Dayiheng  and
      Zhou, Jingren  and
      Lin, Junyang",
    booktitle = "Findings of the Association for Computational Linguistics: ACL 2025",
    month = jul,
    year = "2025",
    address = "Vienna, Austria",
    publisher = "Association for Computational Linguistics",
}

@InProceedings{pmlr-v202-gao23h,
  title = 	 {Scaling Laws for Reward Model Overoptimization},
  author =       {Gao, Leo and Schulman, John and Hilton, Jacob},
  booktitle = 	 {Proceedings of the 40th International Conference on Machine Learning},
  pages = 	 {10835--10866},
  year = 	 {2023},
  volume = 	 {202},
  series = 	 {Proceedings of Machine Learning Research},
  month = 	 {23--29 Jul},
  publisher =    {PMLR},
}

@inproceedings{wang-etal-2025-prejudge,
    title = "Prejudge-Before-Think: Enhancing Large Language Models at Test-Time by Process Prejudge Reasoning",
    author = "Wang, Jianing  and
      Jiang, Jin  and
      Liu, Yang  and
      Zhang, Mengdi  and
      Cai, Xunliang",
    booktitle = "Findings of the Association for Computational Linguistics: EMNLP 2025",
    month = nov,
    year = "2025",
    address = "Suzhou, China",
    publisher = "Association for Computational Linguistics",
}

@article{xiong2025enhancing,
  title={Enhancing Long Chain-of-Thought Reasoning through Multi-Path Plan Aggregation},
  author={Xiong, Siheng and Payani, Ali and Fekri, Faramarz},
  journal={arXiv preprint arXiv:2510.11620},
  year={2025}
}

@article{prabhudesai2025maximizing,
  title={Maximizing confidence alone improves reasoning},
  author={Prabhudesai, Mihir and Chen, Lili and Ippoliti, Alex and Fragkiadaki, Katerina and Liu, Hao and Pathak, Deepak},
  journal={arXiv preprint arXiv:2505.22660},
  year={2025}
}

\appendix

\section{Accuracy-Efficiency Score (AES)}
\label{app:aes}
The AES score \citep{luo2025opruner} compares a tuned model against its corresponding base model to measure the trade-off between accuracy and computational efficiency. Let $\text{L}_{\text{base}}$ and $\text{L}_{\text{model}}$ denote the output lengths, and $\text{Acc}_{\text{base}}$ and $\text{Acc}_{\text{model}}$ denote the accuracies of the base and tuned models, respectively. Defining $\Delta \text{L}=\frac{\text{L}_{\text{base}} - \text{L}_{\text{model}}}{\text{L}_{\text{base}}}$ and $\Delta \text{Acc}=\frac{\text{Acc}_{\text{model}}-\text{Acc}_{\text{base}}}{\text{Acc}_{\text{base}}}$, the AES is calculated as:
\begin{equation*}
\text{AES}
=
\begin{cases}
\alpha \cdot \Delta \text{L} + \beta \cdot \Delta \text{Acc}, & \text{if } \Delta \text{Acc} \ge 0 \\
\alpha \cdot \Delta \text{L} - \gamma \cdot |\Delta \text{Acc}|, & \text{if } \Delta \text{Acc} < 0
\end{cases}
\end{equation*}
where $\alpha > 0$, $\beta > 0$, and $\gamma > 0$. Following the original work, we set $\alpha = 1$, $\beta = 3$, and $\gamma = 5$, with $\gamma > \beta$ to emphasize the penalization of accuracy degradation.

\section{Step Segmentation}
\label{app:step_seg}
The use of double newlines (\texttt{\textbackslash n\textbackslash n}) for step segmentation is grounded in the training data format of the model family we use, rather than being an arbitrary heuristic. From Qwen2.5-Math-1.5B to DeepSeek-R1-Distill-Qwen-1.5B, the SFT stage uses approximately 800K examples \citep{guo2025deepseek}, in which individual reasoning steps are consistently separated by \texttt{\textbackslash n\textbackslash n}. Subsequently, from DeepSeek-R1-Distill-Qwen-1.5B to DeepScaleR-1.5B-Preview, the post-training dataset (DeepScaleR-Preview-Dataset, which we also use) has the same format of separating reasoning steps with \texttt{\textbackslash n\textbackslash n}. Therefore, our segmentation strategy aligns with the formatting pattern embedded throughout both the SFT and RL training stages, rather than introducing a new structural assumption. In addition, prior work \citep{zhang-etal-2025-lessons, wang-etal-2025-prejudge, xiong2025enhancing} also segments reasoning steps using double newlines (\texttt{\textbackslash n\textbackslash n}) naturally, motivated by pretraining priors and the inherent structure of reasoning tasks.

Importantly, SAS operates at the granularity of complete reasoning steps, including the trailing double newline delimiter (\texttt{\textbackslash n\textbackslash n}), and assigns advantages to all tokens within each step collectively. As a result, SAS does not alter the step formatting pattern. Empirically, we observe that the model preserves the original step format during training. Moreover, SAS is not intrinsically tied to the \texttt{\textbackslash n\textbackslash n} delimiter: if a different model family adopts another structured reasoning format, step boundaries can be defined accordingly without modifying the core algorithm.

\section{Licenses}
Datasets are released under the following licenses:
\begin{itemize}
    \item DeepScaleR-Preview-Dataset: MIT license
    \item AIME24: Apache-2.0 license
    \item AIME25: Apache-2.0 license
    \item MATH: MIT license
    \item AMC: Apache-2.0 license
    \item OlympiadBench: MIT license
    \item GPQA: MIT license
    \item LSAT: MIT license
    \item MMLU: MIT license
\end{itemize}
The models we use have the following licenses:
\begin{itemize}
    \item \deepscaler{}: MIT license
    \item L1-Max: MIT license
    \item LAPO-I: Apache-2.0 license
    \item ThinkPrune-4k:Apache-2.0 license
\end{itemize}

\end{document}